# EVALUATING THE IMPACT OF CONVOLUTIONAL NEURAL NETWORK LAYER DEPTH ON THE ENHANCEMENT OF INERTIAL NAVIGATION SYSTEM SOLUTIONS

Mohammed AFTATAH and Khalid ZEBBARA

IMISR Laboratory, Applied Sciences Faculty Ait Melloul, Ibn Zohr University, Agadir, Morocco

## ABSTRACT

*Secure navigation is pivotal for several applications including autonomous vehicles, robotics, and aviation. The inertial navigation system estimates position, velocity, and attitude through dead reckoning especially when external references like GPS are unavailable. However, the three accelerometers and three gyroscopes that compose the system are exposed to various types of errors including bias errors, scale factor errors, and noise, which can significantly degrade the accuracy of navigation constituting also a key vulnerability of this system. This work aims to adopt a supervised convolutional neural network (ConvNet) to address this vulnerability inherent in inertial navigation systems. In addition to this, this paper evaluates the impact of the ConvNet layer's depth on the accuracy of these corrections. This evaluation aims to determine the optimal layer configuration maximizing the effectiveness of error correction in INS (Inertial Navigation System) leading to precise navigation solutions.*

## 1. INTRODUCTION

Inertial Navigation Systems (INS) come in various types, each suited for different applications based on their accuracy, cost, and technological sophistication. Here are the primary types of INS: fiber optic-based gyroscope, Ring laser-based gyroscope, and gyroscopes of the mechanical type, and MEMS (Micro Electro Mechanical Systems) use tiny integrated circuits incorporating accelerometers and gyroscopes. The raw accelerations across the three axes (x, y, and z) are measured by the three-axis accelerometers (axial acceleration), while angular velocity (rotation) is measured by the three-axis gyroscopes around these axes [1]. These systems are much cheaper and smaller than traditional gyroscopic systems, making them ideal for commercial applications like smartphones and drones [2]. However, they are generally less accurate over long periods.

The loss of precision over extended periods for the MEMS-INS type arises from several error sources including scale factor, noise, and bias, which significantly impact their performance. Bias, a constant error that can drift over time, leads to cumulative inaccuracies in navigation outputs [3]. Scale factor errors, which are inconsistencies in the sensor's response relative to the actual motion, cause deviations proportional to the true movement. Noise, comprising both random and systematic variations, further degrades measurement precision, complicating accurate navigation [4]. Recently, many groups of researchers have investigated several neural networks to address these issues, including RNNs (Recurrent Nets), DRL (Deep learning-based reinforcement), and ConvNets [5]. Furthermore, using additional layers enables neural networks to better model and learn the complex relationships between the true motion of the system and the raw measurements of the inertial sensors [6].

The approach proposed by the authors of [7], used a ConvNet (Convolutional Network) with the aim of de-noising inertial measurements. Under constant specific forces and rotations, the researchers gathered measurements from two distinct IMUs: a low-cost IMU and a high-grade IMU. The algorithm used the high-grade IMU dataset to learn patterns and features in the low-cost IMU measurements, ultimately generating a model based on ConvNet. The results from their experiment demonstrate that ConvNet can effectively reduce sensor errors and perform the accuracy. However, the authors have applied the methodology in a simulation-world navigation setup. Furthermore, this methodology does not conclusively show how deep learning can reduce error drifts for inertial sensors in real-world sensors. In [8], the authors employed SRU-RNN for de-noising MEMS gyroscope signals, achieving substantial improvements in reducing noise, bias instability, and angle random walk.

OriNet is a method detailed in [9], which inputs to the network of memory cells (Long Short-Term Memory) the signals from a 3D gyroscope to produce corrected gyroscope signals. A loss function was defined and minimized to measure the difference between the estimated and actual orientations, training the LSTM model. OriNet was evaluated using a data pool from a public drone, which demonstrated an approximate 80% enhancement in attitude performance. In [10], the reinforcement-learning algorithms are employed to identify optimal parameters for inertial calibration algorithms.

In 2019, Zhu et al. [11] introduced a NAS-RNN model to suppress noise in MEMS gyroscope data, demonstrating significant improvements over traditional LSTM-RNN methods, with further decreases in standard deviation values and attitude errors. The authors of [12] described an approach, where a gyroscope is calibrated using a ConvNet, resulting in high accuracy in attitude estimation. Wang et al. proposed in their work [13] a hybrid method integrating CNN-LSTM and PSO-SVM to address temperature-induced errors in MEMS gyroscopes, resulting in a significant reduction of temperature drift and improved gyroscope accuracy.

Moreover, a review paper by Podder et al. [14] provided a comprehensive analysis of artificial intelligence applications in MEMS sensors, underscoring methods such as CNN, RNN, LSTM, and ANN for process optimization, signal de-noising, and error correction, highlighting the potential and effectiveness of artificial intelligence in performing MEMS sensors.

Numerous researchers have employed ConvNet and other AI (Artificial Intelligence) algorithms to mitigate errors in inertial navigation systems. While these approaches have shown promise in reducing such errors, the impact of the depth of layers within these networks has not been thoroughly explored. This gap in the literature suggests that further investigation into the depth of ConvNet layers could provide significant insights and potentially lead to more optimized solutions for addressing INS errors.

Table 1. A summarized table of existing methods for machine-learning-based correction of inertial sensors.

| Authors | Year | Model | Learning | Target sensors of correction |
|---|---|---|---|---|
| Chen et al. [7] | 2018 | ConvNet | Supervised | Accelerometer, gyroscope |
| Jiang et al. [8] | 2018 | Simple Recurrent Unit RNN (SRU-RNN) | Supervised | MEMS Gyroscope |
| Esfahani et al. [9] | 2019 | Orinet | Supervised | Gyroscope |
| Nobre and Heckman [10] | 2019 | Reinforcement learning algorithms | Unsupervised | Optimal parameters for inertial calibration |
| Zhu et al. [11] | 2019 | NAS-RNN | Supervised | MEMS Gyroscope |
| Brossard et al. [12] | 2020 | ConvNet | Supervised | IMU Gyroscope |
| Wang and Cao [13] | 2022 | CNN-LSTM and PSO-SVM | Supervised | MEMS Gyroscope |
| Podder et al. [14] | 2023 | Artificial intelligence Applications | Supervised | MEMS-Based Sensors |

This paper can be summarized by its primary objectives as follows:

- ✓ Modeling INS sensors from a reference trajectory;
- ✓ Modeling the main errors affecting accelerometers and gyroscopes, specifically focusing on bias, scale factor, and noise;
- ✓ Developing a CovNet algorithm to mitigate the errors of the MEMS sensors including both accelerometers and gyroscopes;
- ✓ Exploring the adaptability of the CNN architecture through the manipulation of the number of layers, assessing its impact on error mitigation;
- ✓ Comparison of the performance of the INS estimated positions before and following the use of the proposed algorithm.

Six distinct sections are written to organize this paper: Firstly, Section 1 introduces this work. Secondly, Section 2 contains a review of the literature, an outline of the primary contributions of this paper, an explanation of INS, a discussion of the INS sensors' modeling and their inherent errors, and an introduction to the ConvNets algorithm. Thirdly, the methodology employed to implement the ConvNets algorithm is explained in Section 3 details. Fourthly, Section 4 details the experiment. Additionally, Section 5 presents the findings in terms of the effectiveness of the CNN algorithm in reducing errors in the output of the MEMS-INS. Finally, Section 6 provides the conclusion of the paper by synthesizing the findings and discussing the future work planned by our group.

## 2. BACKGROUND INFORMATION

### 2.1. Introduction To Inertial Navigation

An Inertial Navigation System operates independently (qualified as an autonomous navigation system) and computes through a process known as dead reckoning the velocity, position, and attitude of a mobile without the need for external references. This system utilizes six internal sensors, three gyroscopes, and three accelerometers, respectively measuring raw angular velocities and linear accelerations. These data from the internal sensors are integrated twice over time, enabling the system to estimate with high accuracy the trajectory of the object.

Two well-known types of INS are Micro-Electro-Mechanical Systems INS (MEMS-INS) and Gimbaled INS. This paper focused on MEMS-INS due to many advantages including small dimensions, low cost, and power efficiency, turning them into a perfect choice for an extensive range of applications.

### 2.1.1. Importance of MEMS-INS

The importance of MEMS-INS is evaluated by the extensive use of these sensors in multiple industries. This significance is justified by several key factors:

- Firstly, the ability to operate independently of external signals such as GPS;
- Secondly, the deliverance of continuous navigation ;
- Thirdly, the high-speed computation is ideal for real-time processing and output;
- Fourthly, the resistance to jamming and spoofing attacks;
- Finally, the providence of a wide range of information including velocity, position, and attitude in 3D.

### 2.1.2. Mems-Ins Mechanization

MEMS-INS Mechanization is a set of INS sensors' equations employed to derive navigation outputs, including the mobile's position, velocity, and attitude, from the measurements of inertial sensors, specifically the accelerations, and rotations. In this study, two references are frequently used: the ENU frame (East, North, Up) and the chassis frame (body frame). Upcoming, the dynamic equations are expressed in the ENU frame. Additionally, the ENU system will be referred to as the navigation frame. The chassis frame is defined at the center of the inertial system. These equations are given in (1), (2), and (3) [15, 16, 17]:

$$\dot{r}^{LLA} = D v^n \quad (1)$$

$$\dot{v}^n = C_b^n f_{ib}^b - (2\omega_{ie}^n + \omega_{en}^n) \times v^n + g^n \quad (2)$$

$$\dot{C}_b^n = C_b^n (S(\omega_{nb}^b)) \quad (3)$$

Here, $r^n = \begin{bmatrix} \phi & \lambda & \phi \end{bmatrix}^T$ refers to the latitude, longitude, and altitude in the ENU frame,

$v^n = \begin{bmatrix} v_e & v_n & v_u \end{bmatrix}^T$ presents the linear velocity,

$f_{ib}^b$ are the accelerometer's raw measurements in the body frame,

$C_b^n$ and $C_n^b$ are used to convert values from the chassis frame to the ENU coordinate system and to transform values from the ENU coordinate system to the chassis frame, respectively. They depend on three angles yaw, pitch, and roll ($\psi$, $\theta$, and $\phi$: Euler angles).

$\omega_{nb}^b$ is the angular velocities directly measured by the gyroscopes about the three axes of the INS after removing $\omega_{ie}^b$ and $\omega_{en}^b$, $\omega_{ie}^n$ is the expression in the ENU frame of the Earth's rotation rate,

$\omega_{en}^n$ is the ENU frame's rotation rate vector relative to the Earth,

$g^n$ presents the vector of the gravity,

D presents a matrix to transform values from the navigation frame to the LLA (latitude, longitude, and altitude) frame as given in (4).

$$D = \begin{pmatrix} 0 & \dfrac{1}{R_m + h} & 0 \\[2mm] \dfrac{1}{(R_n + h)\cos\phi} & 0 & 0 \\[2mm] 0 & 0 & 1 \end{pmatrix}$$

(4)

Where,

$R_m$ the radii of curvature in the meridian,

$R_n$ is the prime vertical,

$e^2 = \dfrac{a^2 - b^2}{a^2}$ is the Earth's eccentricity,

$a$ is the Earth's semi-major axis,

$b$ is the Earth's semi-minor axis,

Then, the dynamic equations in the ENU frame can be expressed as:
$$\dot{P}^n = V^n$$

Where, $P^n = \begin{bmatrix} P_e & P_n & P_u \end{bmatrix}^T$ refers to the vector of 3D positions.

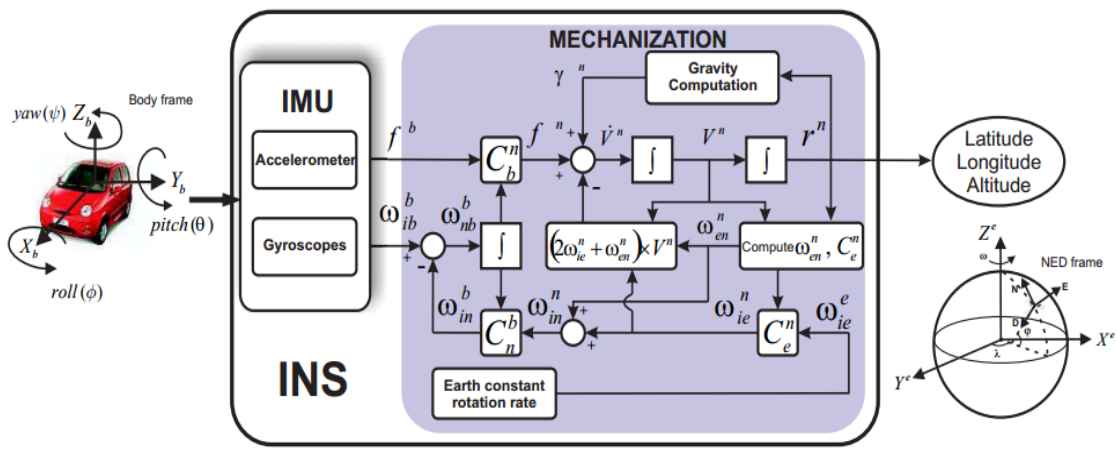

Figure 1.  The mechanization of MEMS-INS, kindly taken form [18, 19]

### 2.1.3. Mems-Ins Inherent Errors

Although the INS can provide real-time navigation data, even in environments where external references are unreachable. However, INS sensors are exposed to many forms of errors that can significantly affect their consistency and accuracy. These errors in INS sensors include biases, scale factor errors, and noise, all of which affect the accelerometers and gyroscope measurements.

Bias errors are constant or slowly varying added to the true value, causing a drift in both angular rate and acceleration measurements. Additionally, scale factor errors result from the deviation of the sensor's sensitivity from its ideal value. Furthermore, noise in INS sensors, including random walk and white noise, complicates the readings even more.
Taking into consideration all these types of errors, the measured acceleration and angular rate can be expressed as (5)(6):

$$\omega_{measured} = (1 + SF_{gyroscope}) * \omega_{true} + b_{gyroscope} + \eta_{gyroscope} \qquad (5)$$

$$a_{measured} = (1 + SF_{accelerometer}) * a_{true} + b_{accelerometer} + \eta_{accelerometer} \qquad (6)$$

Where,

$a_{true}$ and $\omega_{true}$ are the true acceleration and angular rates, while $\eta_{accelerometer}$ and $\eta_{gyroscope}$ represent noise in the measurements,

$b_{accelerometer}$ and $b_{gyroscope}$ represent the bias in the accelerometer and gyroscope respectively,

$SF_{accelerometer}$ and $SF_{gyroscope}$ the scale factor errors for accelerometers and gyroscopes, respectively.

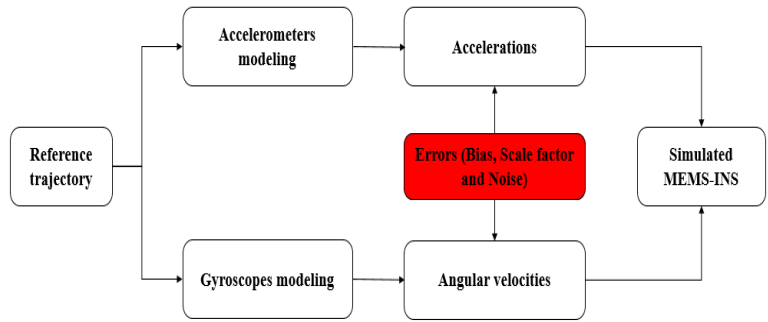

Figure 2. MEMS-INS modeling

## 2.2. ConvNet

ConvNet (Convolutional Neural Network, simply Convolutional Network) is considered as a class of deep neural networks specializing in dealing with grid-structured data, such as visual imagery. However, they have recently been adapted for sequential data tasks like speech recognition and error correction of inertial navigation systems. Subsequently, it provides several advantages, including the ability to handle large volumes of data and automatically detect relevant features without manual intervention. This makes them particularly suitable for detecting features and patterns of INS inherent errors, which traditional methods might not effectively treat.

A CNN typically comprises three key layers: a conv layer, a pool layer, and an FC (fully connected) layer. Its main structure is illustrated in Figure 3 [20, 21, 22].

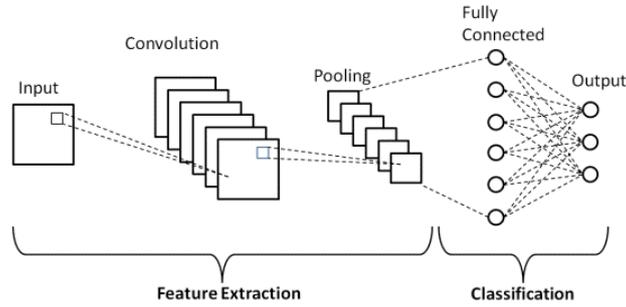

Figure 3. ConvNet's main structure

### 2.2.1. Convolution Layer

The convolutional layer starts by processing the INS estimated positions. These calculated positions contain errors due to three main factors such as bias and sensor noise. The role of this layer is to capture spatial characteristics and patterns from the INS estimates. The convolution is expressed as (7) [23, 24]:

$$h_{ij}^{(1)} = \sigma(\sum_{M=0}^{M-1}\sum_{N=0}^{N-1} x_{(i+m)(j+n)}^{(0)}.\omega_{mn}^{(1)} + b^{(1)}) \qquad (7)$$

Where $x_{(i+m)(j+n)}^{(0)}$ represents the input estimated position data, $\omega_{mn}^{(1)}$ are the weights of the convolutional kernel in the first layer, $b^{(1)}$ refers to the bias term, and σ presents the function of activation.

### 2.2.2. Pooling Layer

After processing the estimated positions by the convolutional layer to extract spatial features, the representation is minimized in spatial dimensions by the pooling layer while preserving the most relevant information, which decreases the necessary total of computation and weights. To capture the most significant features detected by the preceding layer, the Max pooling operation is given by (8) [25]:

$$p_{ij}^{(1)} = \max_{(m,n)\in P(i,j)} h_{mn}^{(1)} \qquad (8)$$

Here, $h_{mn}^{(1)}$ is the output from the convolutional layer at position, P(i,j) represents the pooling region around position, $p_{ij}^{(1)}$ is the pooled output.

### 2.2.3. Fc Layer

After the two layers of convolution and pooling, comes the role of the FC layer (Or the dense layer), which is used in artificial neural networks to connect each neuron or node from the preceding layer to each neuron of the present layer. In this layer, the input vector is linearly converted by the node using a matrix of weights. Then, a non-linear conversion is applied to the result through a non-linear activation function. The Equation for a non-linear transformation for a FC layer is defined as (9) [25]:

$$y_k^{(2)} = \sigma(\sum_j \omega_{kj}^{(2)} p_j^{(1)} + b_k^{(2)})$$                    (9)

Where $p_j^{(1)}$ represents the output from the pooling layer, $\omega_{kj}^{(2)}$ are the weights matrix, $b_k^{(2)}$ is the bias term, $y_k^{(2)}$ denotes the output of the FC layer after passing through a function of activation σ introducing the non-linearity.

### 2.2.4. Activation Function

Several activation functions are present in the literature and are used with ConvNets, such as tanh, sigmoid, and ReLU. However, ReLU is the most commonly used due to its numerous advantages. ReLU is more expressive with the non-linear model, which allows the network to deal with complex features and functions. Simplicity and efficiency are the main advantages of ReLu in terms of overcoming the gradient problem that can occur when based on sigmoid or tanh as activation functions. Its output equals directly the input if the input is positive; otherwise, it equals zero. The ReLu function is given by (10) [26, 27]:

$$\mathrm{Re}\,Lu(x) = \max(0, x)$$                    (10)

Where $x$ presents ReLU's function input.

## 3. METHODOLOGY

The accuracy of the MEMS-INS navigation solution relies on several elements, as discussed in the previous section, and is significantly influenced by the quality and the grade of the accelerometers and gyroscopes and the capability to redress its inherent errors. The ability of the machine learning-based CNN algorithm to mitigate the inertial sensor's errors by training the estimated positions of the INS using the positions from the reference trajectory is the primary objective of this work.

This study investigates diverse depths of CNN layers to perform the accuracy of the MEMS-INS navigation solution. We will evaluate multiple CNN architectures with varying depths to identify the most effective model for correcting INS errors. By systematically comparing the performance of these models, we aim to determine the optimal depth that minimizes these errors.

The proposed machine learning-based CNN algorithm comprises two distinct phases: firstly, the training phase followed by the testing phase secondly. The first phase is detailed in Figure 4.

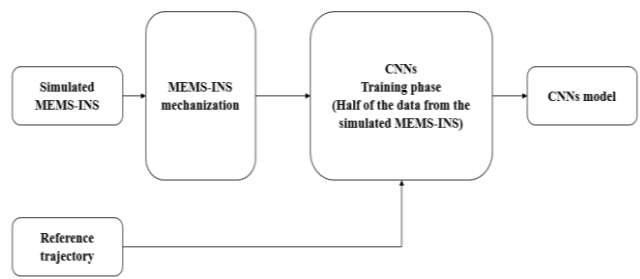

Figure 4. The block diagram representing the first phase (training phase) of machine learning-based CNNs

The proposed algorithm machine learning–based–CNNs for MEMS-INS navigation solution improvement is a process of the following steps:

**Input:** Result of mechanization of three gyroscopes and three accelerometers in terms of INS 3D estimated positions and the reference position.

**Step 1:** Prepare and set the machine learning-based-CNNs configuration (input data, output data, number of layers, types of layers, activation function, and epochs/iterations).

**Step 2:** Consists of the training phase.

**Step 3:** Produce the machine learning-based-CNNs model.

**Step 4:** Includes the testing phase (evaluate and apply the machine learning-based-CNNs on the residual data).

**Step 5:** Examine the output of the machine learning-based-CNNs (improved INS positions).

**Step 6:** Calculate the percentage of enhancement caused by the proposed model using the comparison between the reference positions and the machine learning-based-CNNs values (Root Mean Square Error given by (11)) [28, 29].

$$RMSE = \sqrt{\frac{1}{n}\sum_n (P_{n,ref} - P_{n,CNNs})^2} \qquad (11)$$

Where, $P_{n,CNNs}$ and $P_{n,ref}$ are trained positions and reference positions, respectively.

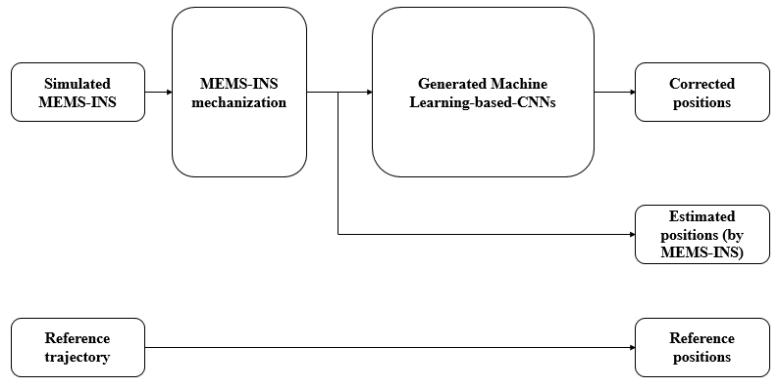

Figure 5. The block diagram detailing the second phase (testing phase) of machine learning-based-CNNs where the generated model is applied to the estimated positions by MEMS-INS and comparing the generated values with the reference once.

The algorithm detailed below will be tested on the INS estimated positions data with three variants: a superficial ConvNet, a medium-depth ConvNet, and a deep ConvNet. These variants are designed to explore the impact of network depth on performance. The configurations of these three variants are detailed in Table 2.

Table 2. The configurations of superficial ConvNet, medium-depth ConvNet, and deep ConvNet

|  | Superficial ConvNet | Medium-depth ConvNet | Deep ConvNet |
|---|---|---|---|
| **Number of layers** | 2 Conv + 2 Pooling + 1 Fully Connected | 4 Conv + 2 Pooling + 2 Fully Connected | 8 Conv + 4 Pooling + 3 Fully Connected |
| **Configuration** | Conv Layer 1: 32 filters, kernel size 3x3, ReLU activation Pooling Layer 1: Max pooling, pool size 2x2 Conv Layer 2: 64 filters, kernel size 3x3, ReLU activation Pooling Layer 2: Max pooling, pool size 2x2 Fully Connected Layer 1: 128 units, ReLU activation Output Layer: Softmax activation (for classification) | Conv Layer 1: 32 filters, kernel size 3x3, ReLU activation Conv Layer 2: 64 filters, kernel size 3x3, ReLU activation Pooling Layer 1: Max pooling, pool size 2x2 Conv Layer 3: 128 filters, kernel size 3x3, ReLU activation Conv Layer 4: 256 filters, kernel size 3x3, ReLU activation Pooling Layer 2: Max pooling, pool size 2x2 Fully Connected Layer 1: 512 units, ReLU activation Fully Connected Layer 2: 256 units, ReLU activation Output Layer: Softmax activation (for classification) | Conv Layer 1: 32 filters, kernel size 3x3, ReLU activation Conv Layer 2: 64 filters, kernel size 3x3, ReLU activation Pooling Layer 1: Max pooling, pool size 2x2 Conv Layer 3: 128 filters, kernel size 3x3, ReLU activation Conv Layer 4: 256 filters, kernel size 3x3, ReLU activation Pooling Layer 2: Max pooling, pool size 2x2 Conv Layer 5: 512 filters, kernel size 3x3, ReLU activation Conv Layer 6: 512 filters, kernel size 3x3, ReLU activation Pooling Layer 3: Max pooling, pool size 2x2 Conv Layer 7: 1024 filters, kernel size 3x3, ReLU activation Conv Layer 8: 1024 filters, kernel size 3x3, ReLU activation Pooling Layer 4: Max pooling, pool size 2x2 Fully Connected Layer 1: 1024 units, ReLU activation Fully Connected Layer 2: 512 units, ReLU activation Fully Connected Layer 3: 256 units, ReLU activation Output Layer: Softmax activation (for classification) |

## 4. EXPERIMENTS

### 4.1. Experimental Setup

Verification of the proposed machine-learning model's effectiveness was conducted using simulated MEMS-INS data instead of real MEMS-INS sensors. The experimental work was performed within a simulation environment using Matlab. The reasons behind this choice come from enabling tests in various scenarios without physical investment and the flexibility and scalability they offer.

From a reference trajectory, we modeled the three accelerometers and three gyroscopes that compose an INS system to generate raw INS measurements, including specific forces and angular velocities. Bias, scale factor, and noise, as inherent INS errors, were then introduced into the data.

These errors were added with specific values commonly affecting the low-grade MEMS-IMU sensors. This method permits to creation of a controlled environment for testing the proposed machine-learning model's performance under conditions that mimic real-world sensor inaccuracies. The performance characteristics of the simulated MEMS-IMU are presented in Table 3 below.

Table 3. Performance characteristics of the simulated MEMS-IMU and real MEMS-IMU

| MEMS-IMU | Real IMU | Simulated IMU |
|---|---|---|
| Output data rate | From 100 Hz to 1 kHz | 100 Hz |
| Warm-up time | around 1 to 5 seconds | 5 seconds |
| Gyroscopes | | |
| Scale factor | 1% to 5% | 5% |
| Bias | 10° to 100°/hour | 100°/hour |
| Noise | 0.01 to 0.1°/s/√Hz | 0.1°/s/√Hz |
| Accelerometers | | |
| Scale factor | 0.1% to 1% | 1% |
| Bias | 100 to 1,000 µg | 1,000 µg |
| Noise | 100 to 500 µg/√Hz | 500 /√Hz |

## 4.2. Simulation Environment

The simulated trajectory, developed within a MATLAB simulation environment, includes both curved paths and straights to effectively emulate real scenarios. Indeed, curved segments introduce more complex changes in attitude and velocity, while straight paths represent simple linear motion. The following figures (Figure 6 and Figure 7) illustrate the initial simulated trajectory in both 2D and 3D, highlighting the challenges posed by different motion dynamics.

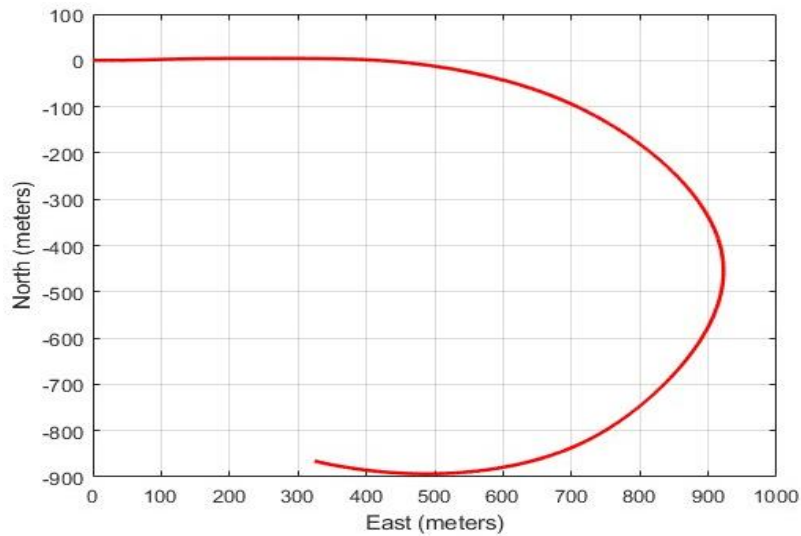

Figure 6. Overview of the 2D simulated trajectory

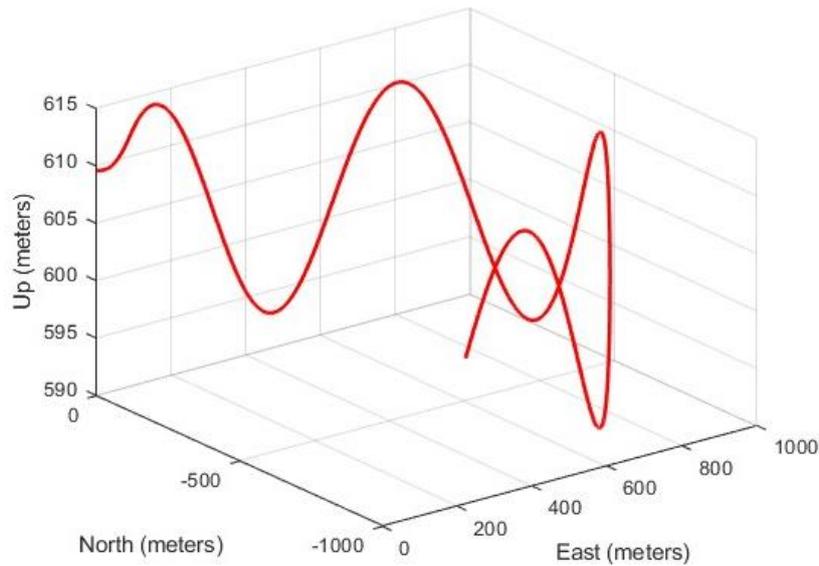

Figure 7.  Summary of the 3D simulated trajectory

## 5. RESULTS & DISCUSSION

The simulated trajectory, accurately reflecting real-world conditions, was employed to evaluate our proposed ConvNet variants within the MATLAB environment. At the frequency of 1 Hz, this trajectory consists of 3000 points, corresponding to a duration of 3000 seconds. This duration translates to 50 minutes, providing a substantial dataset for assessing the performance and robustness of our ConvNet models in various scenarios. By utilizing this extensive trajectory data, the error correction algorithms were thoroughly tested and validated, ensuring their practical applicability and reliability.

The three configurations of the ConvNet were applied to the erroneous data in a two-phase process. In the first stage, 50% of the position data from the reference trajectory was utilized to generate the three machine-learning-based ConvNet models. Following this, the three proposed models were applied to the remaining INS erroneous data set to measure their performance. This approach allowed for an initial training phase using a substantial portion of the accurate data, ensuring the models were well-calibrated before evaluating their effectiveness on the unseen, erroneous portion of the dataset.

In all upcoming figures, the following color scheme was utilized to distinguish between the different datasets: red was used to represent the reference trajectory. The data processed by the first ConvNet configuration is shown in blue. The second ConvNet configuration's results are depicted in green. Finally, the data corrected by the third ConvNet configuration is displayed in magenta.

The East, North, and Up position components are compared in Figures 8-10. By analyzing these figures, the obtained position components using the proposed ConvNet models were substantially closer to the reference position components. Furthermore, the findings indicate that the accuracy of the position estimation improved with the increasing depth of the ConvNet architecture. Specifically, deeper ConvNet models resulted in positional outputs that were more closely aligned with the reference data. This suggests that deeper neural network structures have enhanced capability in mitigating INS errors and refining positional accuracy, thereby validating the efficacy of employing deeper ConvNet configurations for improved inertial navigation performance.

However, it was also observed that the computational time required is proportional to the network depth. This increase in computation time underscores the substantial use of CPU (Central Processing Unit) resources necessary to handle deeper network structures. Consequently, while deeper ConvNet configurations offer superior accuracy, they also demand greater computational power and resources, highlighting a trade-off between performance accuracy and computational efficiency. This balance must be carefully managed to optimize both the effectiveness and efficiency of the INS error correction process.

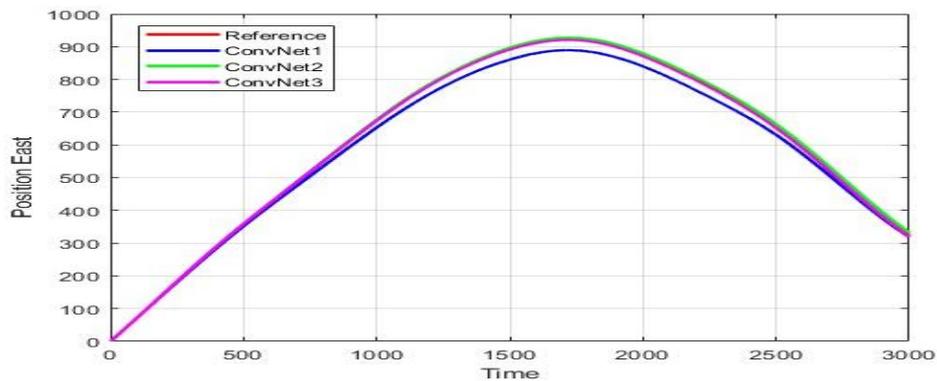

(a)

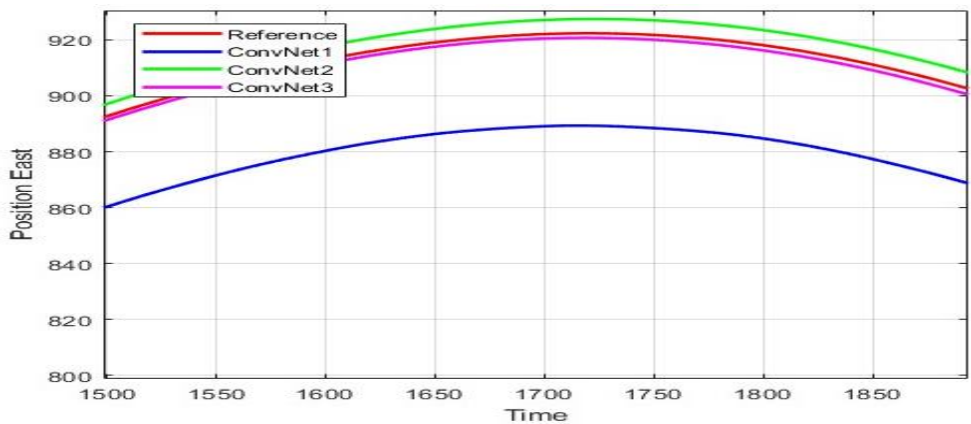

(b)

Figure 8. Evaluation of the three ConvNet models targeting the east component of positional data (a) & zoomed-in view (b)

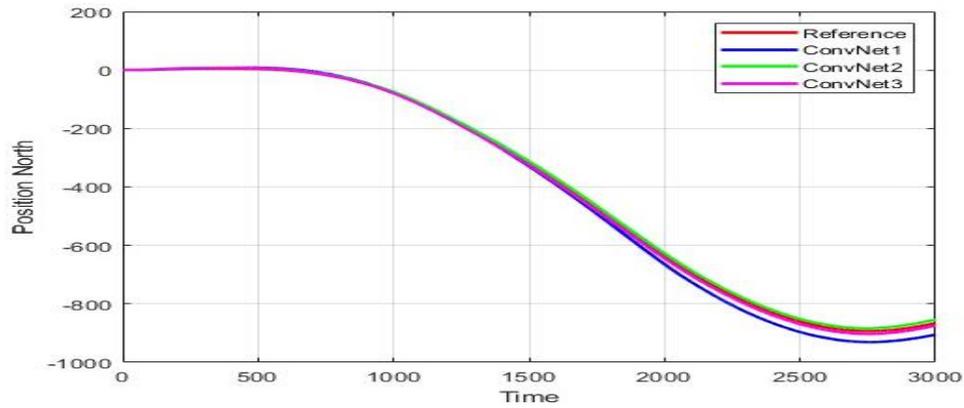

(a)

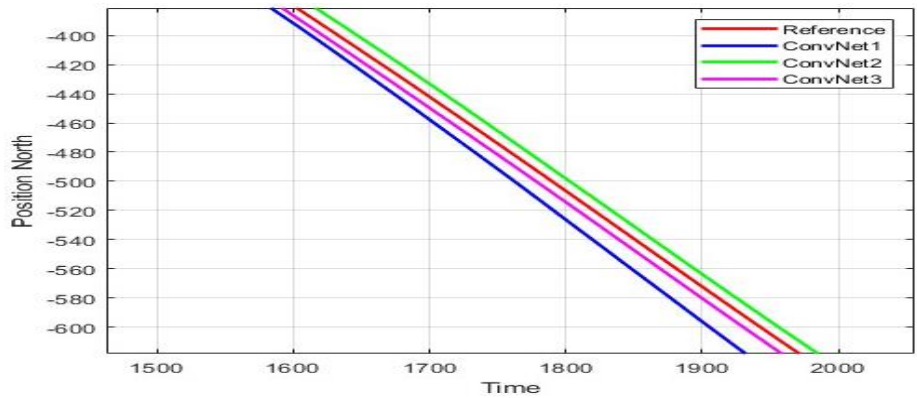

(b)

Figure 9. Evaluation of the three ConvNet models targeting the north component of positional data (a) & zoomed-in view (b)

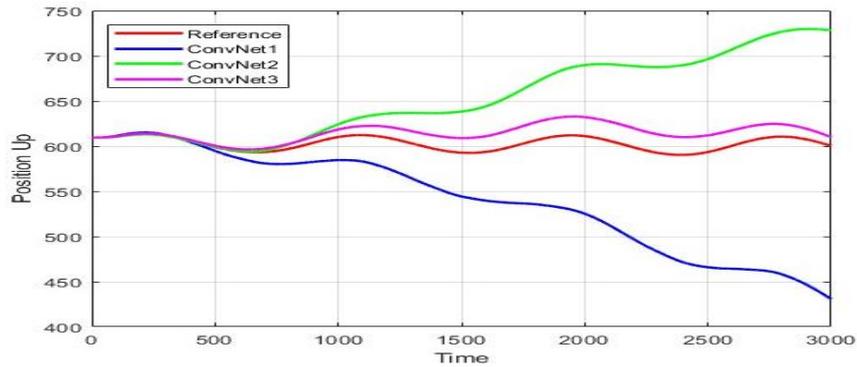

Figure 10. Evaluation of the three ConvNet models targeting the up component of positional data

The overall 2D and 3D trajectory comparisons are presented in Figure 11. As demonstrated by the illustrations, the machine learning-based ConvNet technique effectively mitigates the inherent errors of MEMS-INS, leading to an improved INS navigation solution. Furthermore, the ConvNet3 architecture produces the most significant and relevant results. The obtained trajectory by the ConvNet3 closely tracked the reference path, particularly through the curve segments.

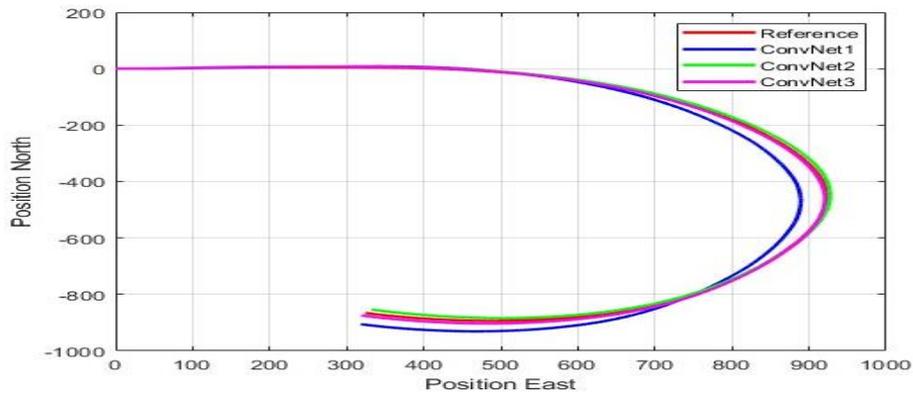

(a)

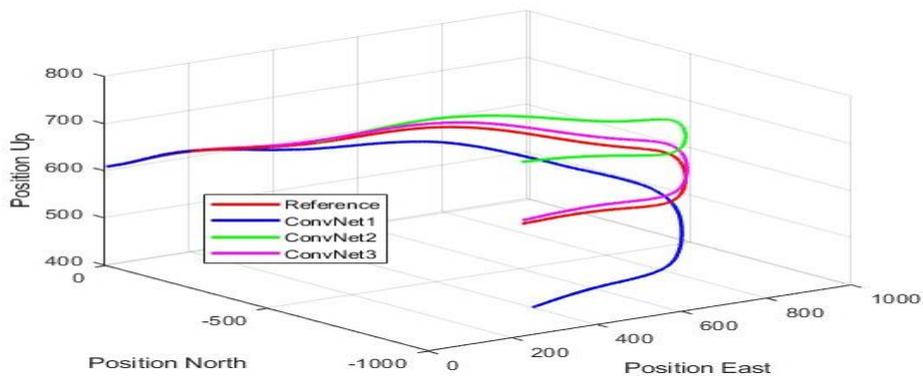

(b)

Figure 11.  Results after applying the three ConvNet configurations to the trajectory in both 2D (a) and 3D

(b)

Table 4 provides the RMSE (Root Mean Square Error) analysis of the INS solution's position components. This analysis underscores the improved accuracy achieved with the ConvNet3 method, outperforming the ConvNet1 and ConvNet2 architectures in reducing INS errors and enhancing trajectory accuracy. However, the RMSE in the up direction remains high even with the deepest architecture provided by the ConvNet3, due to gravity fluctuations caused by variations in the Earth's gravitational field.

Table 4.  RMSE analysis of East, North, and Up position components for the ConvNet variants

| Position RMSE | Superficial ConvNet | Medium depth ConvNet | Deep ConvNet |
|---|---|---|---|
| East (m) | 23.63 | 5.08 | 2.39 |
| North (m) | 22.10 | 6.93 | 6.31 |
| Up (m) | 81.50 | 66.47 | 14.10 |

The following table (Table 5) summarizes the findings by comparing the key parameters of performance, including CPU utilization, accuracy, and RMSE's average, for the three ConvNet variants. ConvNet3 demonstrates superior accuracy and lower RMSE but requires higher computational time and resource utilization. On the other hand, ConvNet1 and ConvNet2 provide a more balanced trade-off between accuracy and efficiency.

Table 5.  Summary of the findings by comparing the three ConvNet variants: A Detailed Comparison

| Parameter | Superficial ConvNet | Medium depth ConvNet | Deep ConvNet |
|---|---|---|---|
| Architecture | 2 conv + 2 pooling + 1 FC | 4 conv + 2 pooling + 2 FC | 8 conv + 4 pooling + 3 FC |
| Activation function | ReLu | | |
| Framework | Matlab environment | | |
| CPU Utilization | Low | Medium | High |
| RMSE average over 3D in meters (East, North, and Up) | 42.41 | 26.16 | 7.60 |
| Accuracy | 69.70% | 70.20% | 91.72% |

## 6. CONCLUSIONS & FUTURE WORK

To summarize this paper, our work highlights the fundamental role of layer depth in enhancing the MEMS-INS navigation solution performance within the context of supervised machine learning, specifically relying on the ConvNet model. Using a dataset obtained by modeling and mechanizing the INS system from a reference trajectory that was generated to reflect real scenarios. Three variants of ConvNet were applied to this dataset to improve the accuracy in correcting INS inherent errors. The results demonstrate that ConvNet3 achieves high accuracy and outperforms the two other variants, but it requires more CPU resources. ConvNet 3, the deep convolutional neural network, achieved an accuracy of 91.72%. In comparison, the two other variants did not exceed an accuracy of 80%. On the other hand, ConvNet1 and ConvNet2 offer a better balance between efficiency and accuracy. This study highlights the critical trade-offs between computational resource consumption and model performance, guiding future developments in the field of MEMS-INS error correction.

While mitigating inherent errors in inertial navigation systems (INS) is crucial, it is not sufficient to ensure secure navigation for intelligent systems. To achieve robust navigation, especially in challenging GNSS environments, further steps are necessary. As a next step, we will combine this work with either machine learning techniques or the Kalman Filter to bridge GPS outages. This integration aims to provide a more secure and resilient solution for intelligent systems in GPS-denied environments where their signals are unreliable or unavailable.

## CONFLICTS OF INTEREST

The authors declare no conflict of interest.

### ACKNOWLEDGEMENTS

The contributors of this paper appreciate the cooperation and collaboration of our laboratory members.

## AUTHORS

**Mohammed AFTATAH** obtained a state engineer degree in Network and Telecommunications from ENSA Marrakech, Cadi Ayyad University, Marrakech, Morocco. Presently, he is a trainer at OFPPT in Network, Systems, and Cybersecurity and is actively pursuing his Ph.D. in Artificial Intelligence and its application to secure navigation for intelligent systems.

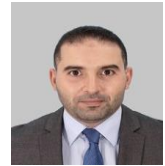

**Dr. Khalid ZEBBARA** earned his Ph.D. in Computer Systems from Ibn Zohr University, Agadir, Morocco. He is currently a Professor at the Faculty of Applied Sciences, Ibn Zohr University, Agadir. In addition, he heads the Imaging, Embedded Systems, and Telecommunications (IMIS) research team at the Faculty of Applied Sciences, Ibn Zohr University, Agadir.

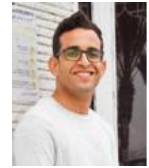